\documentclass[runningheads]{llncs}

\usepackage{eccv}

\usepackage{eccvabbrv}

\usepackage{graphicx}

\usepackage[accsupp]{axessibility}  

\usepackage{booktabs}
\usepackage{bbm}

\usepackage{hyperref}

\usepackage{orcidlink}

\usepackage{rotating}
\usepackage{svg}

\usepackage{pifont}
\newcommand{\cmark}{\ding{51}}%
\newcommand{\xmark}{\ding{55}}%

\newcommand{\hmioum}{\textrm{HM IoU}_{\textrm{multi}}}

\begin{document}

\title{Deterministic Mode Proposals: An Efficient Alternative to Generative Sampling for Ambiguous Segmentation} 

\titlerunning{Mode proposal models}

\author{Sebastian Gerard\orcidlink{0000-0002-5329-8184} \and
Josephine Sullivan\orcidlink{0000-0003-2784-7300}}

\authorrunning{S. Gerard and J. Sullivan}

\institute{KTH Royal Institute of Technology, Stockholm, Sweden\\
\email{\{sgerard, sullivan\}@kth.se}}

\maketitle

\begin{abstract}
Many segmentation tasks, such as medical image segmentation or future state prediction, are inherently ambiguous, meaning that multiple predictions are equally correct. Current methods typically rely on generative models to capture this uncertainty. However, identify the underlying modes of the distribution with these methods is computationally expensive, requiring large numbers of samples and post-hoc clustering. In this paper, we shift the focus from stochastic sampling to the direct generation of likely outcomes. We introduce mode proposal models, a deterministic framework that efficiently produces a fixed-size set of proposal masks in a single forward pass. To handle superfluous proposals, we adapt a confidence mechanism, traditionally used in object detection, to the high-dimensional space of segmentation masks. Our approach significantly reduces inference time while achieving higher ground-truth coverage than existing generative models. Furthermore, we demonstrate that our model can be trained without knowing the full distribution of outcomes, making it applicable to real-world datasets. Finally, we show that by decomposing the velocity field of a pre-trained flow model, we can efficiently estimate prior mode probabilities for our proposals. 
\end{abstract}

\section{Introduction}
\label{sec:intro}

Irreducible (aleatoric) uncertainty is inherent in many prediction tasks. While traditional modeling often focuses on estimating calibrated outcome probabilities, many critical vision tasks face a more fundamental challenge: identifying the set of plausible outcomes within a high-dimensional solution space. For example, predicting wildfire progression through heterogeneous terrain, identifying potentially malignant areas on a lung scan, or forecasting future locations of a child who is temporarily occluded by a parked vehicle, all require dealing with ambiguity where multiple solutions are equally valid. 

Previous work~\cite{kohl_probabilistic_2018,zhang_probabilistic_2022,monteiro_stochastic_2020,de2025flow,rahman_ambiguous_2023,chen_berdiff_2023,zbinden_stochastic_2023,gerard_wildfire_2025,huang_p2sam_2024} has taken a generative perspective, sampling random predictions from learned probability distributions. With this approach, an arbitrary number of predictions can be generated, with good image quality and according to well-calibrated probabilities. However, the downside of this class of approaches is that finding the full set of likely outcomes and their probabilities becomes difficult: Large numbers of samples need to be generated and clustered, incurring high cost.

\begin{figure}
    \centering
    \includegraphics[width=\linewidth]{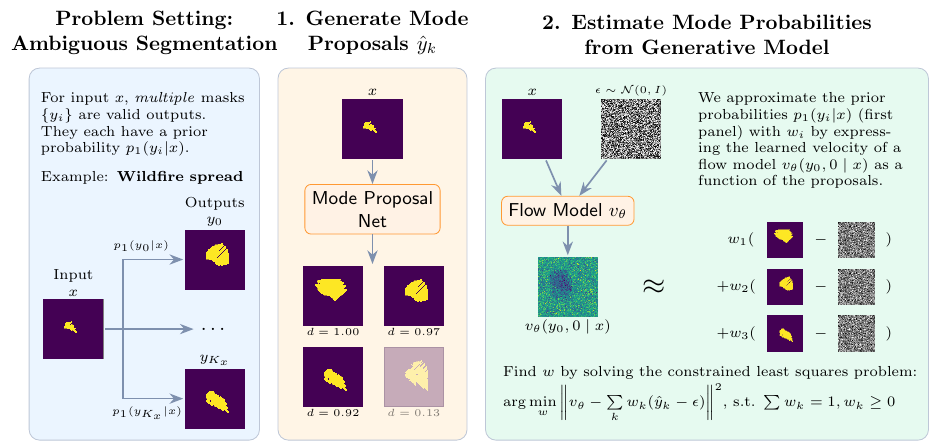}
    \caption{\textbf{Method overview:} In ambiguous segmentation, multiple segmentation masks are valid solutions for a given input, each associated with a prior probability. Our mode proposal model deterministically produces a fixed-size set of proposals and corresponding scores $d\in[0,1]$. To select the most promising candidates, we threshold the selection scores. Finally, we use a separate flow model, trained on the same data, to estimate the prior probabilities of each proposed mode.}
    \label{fig:method-overview}
\end{figure}

In this paper, we take the complimentary approach, focusing on generating the set of plausible outcomes as our primary goal. We argue that achieving this generation efficiently is, for many tasks, more important than modeling the probabilities. For example in safety-related applications (wildfires, medical imaging, autonomous driving), knowing possible scenarios allows for risk mitigation and contingency planning, while knowing the exact probabilities is less important. 

Our \textit{mode proposal model} is a fully deterministic network that outputs a fixed-size set of segmentation masks in a single forward pass. This avoids the stochastic generation of a large set of samples required by generative models, and the expensive multi-step sampling used by diffusion models. Since the number of unique outcomes, or \textit{modes}, is unknown a priori, we assume that we can estimate a safe upper bound on this number. To discard redundant proposals, we draw an analogy to object detection, where large sets of object proposals are filtered down. Finally, we show that we the remaining proposals can be combined with a trained flow model to efficiently estimate the probabilities for each proposal. \cref{fig:method-overview} visualizes the different aspects of our approach.

\paragraph{Our contributions are as follows:}
\begin{itemize}
    \item We introduce the \textit{mode proposal model}, a deterministic model which  efficiently estimates the set of ground truths in ambiguous segmentation tasks with a fixed-size set of proposals. It greatly decreases inference time compared to existing generative models, while achieving a higher coverage. (\cref{tab:main-results})
    \item We transfer the mechanism of \textit{selection scores} from the low-dimensional proposal space of object detection to the high-dimensional proposal space of segmentation tasks, allowing us to filter redundant proposals. (\cref{tab:filter-proposals-all-datasets})
    \item We demonstrate that our approach even works if only individual targets are available during each training step, instead of all targets at once, which is a realistic setting for any future prediction tasks. (\cref{tab:main-results})
    \item We propose a way to decompose the velocity field of a pre-trained flow model to efficiently estimate the mode probabilities, assuming that the modes are sufficiently diverse. (\cref{tab:velocity-decomp-gt-masks,tab:velocity-decomp-filter-proposals})
\end{itemize}

\section{Related Work}

\textbf{Multiple Choice Learning (MCL)}~\cite{guzman-rivera_multiple_2012,lee_stochastic_2016,tian_versatile_2019} assumes each ensemble member to be an expert for a data subset. Deep learning variants~\cite{lee_stochastic_2016,tian_versatile_2019} thus only train the current best-fitting model per input, matching our training approach for the single-label scenario. MCL deals with \textit{input}-space diversity, while we focus on \textit{output}-space diversity, for each given input. Thus, while MCL produces a single sample, we produce multiple proposals, evaluate them with multi-modal datasets, and additionally investigate the fully-labeled scenario. 

The \textbf{mixture of stochastic experts (MoSE)} framework~\cite{gao2023modeling} consists of a shared encoder-decoder structure, branching into light-weight (stochastic) experts, similar to our approach. MoSE learns prior probabilities per mode via an optimal transport loss. While this is conceptually more elegant than ours in the fully-labeled case, the optimal transport loss would collapse in the single-label scenario. In contrast, our approach reliably learns proposals and estimate prior probabilities in the single-label scenario, assuming sufficiently diverse modes. 

Most work on ambiguous segmentation uses \textbf{generative approach}. Early work on the probabilistic U-Net (ProbU-Net)~\cite{kohl_probabilistic_2018} and joint ProbU-Net (JProbU-Net)~\cite{zhang_probabilistic_2022} use conditional variational auto-encoders, generating images based on low-dimensional latent representations, sampled from a learned distribution. To better capture spatial correlations, Stochastic Segmentation Networks (SSN)~\cite{monteiro_stochastic_2020} model the joint distribution over pixels with a low-rank multivariate normal distribution. FlowSSN~\cite{de2025flow} improves upon this, using flows to estimate covariances of arbitrary rank. Many recent methods use the flow-related diffusion models: CIMD~\cite{rahman_ambiguous_2023} combines the ProbU-Net's approach to learning the latent distribution with the diffusion approach to generation. While BerDiff~\cite{chen_berdiff_2023} and CCDM~\cite{zbinden_stochastic_2023} use discrete latent spaces, CIMD and Wildfire Spread Scenarios (WSS)~\cite{gerard_wildfire_2025} threshold Gaussian diffusion. WSS additionally applies the guidance methods PG~\cite{corso_particle_2023} and SPELL~\cite{kirchhof_shielded_2025} to increase the mode coverage. P²SAM~\cite{huang_p2sam_2024} learns a distribution over prompts for the Segment Anything (SAM)~\cite{kirillov_segment_2023} model. 

While such stochastic approaches require many Monte Carlo samples and possibly iterative generation to find the set of unique modes for challenging datasets, our approach uses an \textbf{entirely deterministic} single forward-pass. 


\section{Method}
\label{sec:method}

For an image $x \in \mathcal{X} = \mathbb{R}^{C \times H \times W}$ with $C$ channels and spatial dimension $H \times W$ we want to predict the $0 \leq K_x \leq K$ distinct binary masks consistent with $x$. Inspired by the work in object detection \cite{carion2020end} we view this as a set prediction task. We define a \textit{mode proposal network} that outputs $\hat{K}$ proposal masks $\{y_{xi}, \ldots, y_{x\hat{K}}\}$ with each $y_{xi} \in \{0, 1\}^{H \times W}$, where $\hat{K}$ is chosen such that $K\leq \hat{K}$. To train this network we need to define a loss invariant to permutations of the proposal network's outputs. This is achieved by finding a bipartite matching between the labeled ground truth masks and the predicted masks via the Hungarian matching algorithm. As $K$ will generally be larger than $K_x$, we also simultaneously train an auxiliary network to automatically identify the minimal set of proposed masks corresponding to the $K_x$ distinct modes.

We assume two distinct scenarios for the labeled data available at training: The \textit{fully-labeled} scenario assumes that all the modes are available when $x$ is encountered, while single-label assumes just one of the labeled masks is available:
\begin{align}
  \text{Fully-labeled:}\;& (x, y_{1}, \ldots, y_{K_x})\\
  \text{Single-label:}\;& (x, y)\quad \text{where $y \in \{y_1, \ldots, y_{K_x}\}$}
\end{align}

For real-world future-prediction tasks, the \textit{single-label} scenario is more realistic, as in the real world only one future can be observed at a time. We associate each mask with a prior probability $p_1(y_i \mid x)$ to model that outcomes are observed with different frequencies. In our setup, the model eventually sees all $y_i$ for a given $x$ and thus learns to associate them with each other. For real-world data, we assume that the model may learn to generalize from seeing diverse outputs for \textit{similar} inputs.

We now provide more details for training our \textit{mode proposal network} where we adapt the DETR loss \cite{carion2020end} to our problem. 

\subsection{Mode Proposal Network}

We construct a \textbf{mode proposal network} as a convolutional segmentation network with $K$ heads, each one outputting a binary segmentation mask. Let the function $g_{\psi_0}$, parameterized by $\psi_0$, represent these heads and  $f_{\phi}$ represent the network prior to these heads. The \textit{mode proposal network} then maps input $x$ to $K$ predictions, or \textit{proposals}:
\begin{align}
  g_{\psi_0}(f_{\phi}(x)) = \{\hat{y}_{1}^x, \ldots, \hat{y}_{K}^x\}\; \in \{0,1\}^K.
\end{align}
As $K \geq K_x$, some of the proposals output by the network may be duplicates or have low quality. To identify the best proposals, we follow DETR and produce a selection score for each of the $K$ proposals. More formally, we define one more network head $h_{\psi_1}$ that produces the selection scores $d_{k}^x$:
\begin{align}
    h_{\psi_1}(f_{\phi}(x)) = (d_{1}^x, \ldots, d_{K}^x) \in [0, 1]^K.
\end{align}
Each $d_{k}^x$ can be seen as the probability that the $k$th proposal is the best representative for a distinct mode. Ideally, after training $\sum_{k=1}^K \mathbbm{1}_{[d_{k}^x > 0.5]} = K_x$.

\subsubsection{Training in the fully labeled scenario}

To train, we need to find an optimal bipartite matching between the proposals and the labeled modes. This requires finding the best assignment of the $K_x$ labeled masks to a $K_x$ sized subset of the predicted masks. Let $\pi$ be a $K_x$ partial permutation of integers $1, \dots, K$, then 
\begin{align}
\label{eq:optimal-matching}
\pi^* = \underset{\pi}{\arg\min}\; \frac{1}{K_x} \sum_{i=1}^{K_x} l_{\text{\tiny mask}}(y_{i}, \hat{y}_{\pi(i)}^x)
\end{align}
where we define $l_{\text{\tiny mask}}(y, \hat{y})$ to be 1-IoU score between the binary masks $y$ and $\hat{y}$.
This assignment can be computed efficiently with Hungarian matching.

\paragraph{The proposal mask head loss}
Given the optimal assignment $\pi^*$ we can now define the loss for all the matched mask proposal heads 
\begin{align}
\label{eq:lmask}
\mathcal{L}_{\text{\tiny mask}}(x, \phi, \psi_0) = \frac{1}{K_x} \sum_{i=1}^{K_x} l_{\text{\tiny mask}}(y_{i}, \hat{y}_{\pi^*(i)}^x)
\end{align}

\paragraph{Selection head loss} Now we turn our attention to training the selection head.  Let $\mathcal{J}$ be the indices of the proposals not matched to a labeled mask (\ie {\scriptsize $\mathcal{J} = \{1, \ldots, K\} \backslash \{\pi^*(1), \ldots, \pi^*(K_x)\}$}). We then define the loss to train $h_{\psi_1}$ as:
\begin{align}
  \mathcal{L}_{\text{\tiny select}}(x, \phi, \psi_1) = \frac{1}{K} \left[\sum_{i=1}^{K_x} l_{\text{\tiny select}}(1, d_{\pi^*(i)}^x) + \sum_{j \in \mathcal{J}} l_{\text{\tiny select}}(0, d_{j}^x) \right]
  \label{eqn:select_fully}
\end{align}
where we define $l_{\text{\tiny select}}(t, d)$, with $t \in \{0, 1\}$, to the binary cross-entropy loss. 

\paragraph{The total loss} The complete loss used for training is defined for $\lambda \in \mathbb{R}^+$ as
\begin{align}
  \mathcal{L}_{\text{\tiny total}}(x, \phi, \psi_0, \psi_1) = \mathcal{L}_{\text{\tiny mask}}(x, \phi, \psi_0) + \lambda\, \mathcal{L}_{\text{\tiny select}}(x, \phi, \psi_1)
\end{align}

\subsubsection{Training in the single-label scenario}

Now we assume each training example just has one labeled mask that is $(x, y) \in \mathcal{D} = \mathcal{X} \times \{0, 1\}^{H \times W}$. In this case, if the proposal network produces proposals that correspond to a mode not represented by $y$, we refer to these as unlabeled positive proposals. Because of these unlabeled positives, the selection loss in \cref{eqn:select_fully} cannot be used as is and must be amended to train the selection head. We can make use of existing methods from positive-unlabeled learning \cite{de_comite_positive_1999,letouzey_learning_2000} and in particular the non-negative PU loss \cite{kiryo_positive-unlabeled_2017}. This requires a batch, $\mathcal{B}\subset \mathcal{D}$, of training examples where $|\mathcal{B}| = n_b$.  Before defining the PU loss we need to introduce two separate quantities. The first is the mean loss of all \textit{positive} examples in $\mathcal{B}$ when evaluated against positive (or negative) targets,
\begin{align}
  \mathcal{L}_P(t, \mathcal{B}, \phi, \psi_1) = \frac{1}{n_b} \sum_{(x,y) \in \mathcal{B}}\; l_{\text{\tiny select}}(t, d_{\pi^*_x(1)}^x)
\end{align}
where $t \in \{0, 1\}$ is the target and $\pi^*_x(1)$ is the index of the proposed mask which is closest to the labeled mask $y$. The second quantity is the mean loss of all \textit{unlabeled} examples in the current batch evaluated against negative targets. With $\mathcal{J}_x = \{1, \ldots, K\} \backslash \{\pi^*_x(1)\}$, this loss is defined as:
\begin{align}
  \mathcal{L}_U(\mathcal{B}, \phi, \psi_1) = \frac{1}{n_b(K-1)} \sum_{(x,y) \in \mathcal{B}} \sum_{j \in \mathcal{J}_x} l_{\text{\tiny select}}(0, d_{j}^x)~.
\end{align}
The PU loss that replaces \cref{eqn:select_fully} for single-labeled data is then:
\begin{align}
  \mathcal{L}_{\text{\tiny select}}^{\text{\tiny PU}}(\mathcal{B}, \phi, \psi_1) & = \nonumber \\
  & \hspace*{-15pt}\pi_{p} \mathcal{L}_P(1, \mathcal{B}, \phi, \psi_1) + \max\{0, \mathcal{L}_U(\mathcal{B}, \phi, \psi_1) - \eta_p \mathcal{L}_P(0, \mathcal{B}, \phi, \psi_1)\}~, 
\end{align}
where $\eta_p$ is the prior probability of a proposal being positive. The total loss used for training in the single labeled case is then:
\begin{align}
  \mathcal{L}_{\text{\tiny total}}^{\text{\tiny PU}}(\mathcal{B}, \phi, \psi_0, \psi_1) = \frac{1}{n_b} \sum_{(x,y) \in \mathcal{B}} l_{\text{\tiny mask}}(y, \hat{y}_{\hat{\pi}_x(1)}^x) + \lambda\, \mathcal{L}_{\text{\tiny select}}^{\text{\tiny PU}}(\mathcal{B}, \phi, \psi_1)
\end{align}

\subsection{Estimating Mode Probabilities from Generative Models}
\label{sec:method-estimate-probabilities}

The approach above efficiently produces a set of candidate modes, but does not assign probabilities to the selected proposals. In principle, a generative segmentation diffusion model trained on single-label data can estimate mode frequencies by repeated sampling, but this requires extensive sampling to recover low-probability modes and also clustering to associate samples with a mode.

But a trained diffusion model already implicitly encodes information about the relative prevalence of different modes. We show how to extract these mode probabilities directly, avoiding extensive sampling and clustering. For clarity our derivation uses a flow-matching formulation with Gaussian conditional probability paths~\cite{lipman_flow_2024}. This corresponds to the probability paths used by denoising diffusion models. Therefore our results can be adapted to networks trained under a generative diffusion formulation \cite{lipman_flow_2024,schusterbauer_diff2flow_2025}.

\paragraph{Flow-matching setup}
Following a standard flow matching setup, a segmentation mask $y = y_1$ is generated from an initial Gaussian noise sample $y_0 \sim N(0, I)$, conditioned on image $x$, by solving the following ODE:
\begin{align}
  \frac{d}{dt}\, y_t = v(y_t, t \mid x) \qquad \text{with $t \in [0,1]$},
\end{align}
where $v(y_t, t \mid x)$ is the velocity field. For $0 < t < 1$, $y_t$ is a noised version of the clean mask $y$. The ODE transports $p_0(y_0 \mid x) = N(0, I)$ to the target density $p_1(y \mid x)$ through intermediate distributions $p_t(y_t \mid x)$.

Following Lipman \etal~\cite{lipman_flow_2024}, the velocity field is written as an aggregation over $y$ of the conditional velocity fields $v(y_t, t \mid y)$:
\begin{align}
  v(y_t, t \mid x)
  &\overset{\textcolor{white}{\text{Bayes}}}{=} \int v(y_t, t \mid y)\, p_{1 \mid t}(y \mid y_t, x)\, dy \\
  &\overset{\text{Bayes}}{=} \frac{1}{p_t(y_t \mid x)} 
     \int v(y_t, t \mid y)\, p_{t \mid 1}(y_t \mid y)\, p_1(y \mid x)\, dy.
\label{eq:conditional_velocity_integral}
\end{align}
For fixed $y$, the conditional velocity field defines the ODE $\frac{d}{dt} y_t = v(y_t, t \mid y)$ which transports the noise distribution to that mask via the intermediate distributions $p_{t \mid 1}(y_t \mid y)$.
As in diffusion models, we parameterize
\begin{align}
  y_t = \alpha_t y + \beta_t y_0,
\end{align}
with a differentiable, monotonic noise scheduler satisfying $\alpha_0 = 0$, $\beta_0 = 1$, $\alpha_1 = 1$, and $\beta_1 = 0$, so that $p_{t \mid 1}(y_t \mid y)$ is Gaussian. This definition ensures $v(y_t, t \mid y)$ has a simple analytic form defined by differentiating the interpolation:
\begin{align}
  v(y_t, t \mid y)
  = \left(\dot{\alpha}_t - \frac{\dot{\beta}_t}{\beta_t}\alpha_t \right) y
  + \frac{\dot{\beta}_t}{\beta_t} y_t.
\label{eq:conditional_velocity}
\end{align}

\paragraph{Simple approximation of the segmentation mask distribution}
We assume the distribution $p_{1}(y \mid x)$, conditioned on an input image $x$, is discrete and only supported on the set of proposed modes $\{ \hat{y}_1^x, \ldots, \hat{y}_{K_x}^x \}$ that is
\begin{align}
p_{1}(y \mid x) = \sum_{k=1}^{K_x} w_k^x \,\delta(y - \hat{y}_k^x),
\label{eq:discrete_data_distribution}
\end{align}
where $w_k^x \ge 0$, $\sum_{k=1}^{K_x} w_k^x = 1$, and $\delta(\cdot)$ represents the dirac-delta function. Our objective is to estimate the mode probabilities $w_1^x, \ldots, w_{K_x}^x$.

\paragraph{Simplification of the velocity field}
Substituting the discrete distribution from eqn.~\eqref{eq:discrete_data_distribution} into eqn.~\eqref{eq:conditional_velocity_integral}, the integral simplifies to:
\begin{align}
  v(y_t, t \mid x) 
                    &= \frac{1}{p_{t}(y_t \mid x)} \sum_{k=1}^{K_x} w_k^x\, v(y_t, t \mid \hat{y}_k^x)\, p_{t \mid 1}(y_t \mid \hat{y}_k^x)
\label{eq:velocity_mixture}
\end{align}
The marginal velocity field is therefore a mixture of the mode-induced conditional velocity fields $v(y_t, t \mid \hat{y}_k^x)$. Eqn.~\eqref{eq:velocity_mixture} simplifies even further if we assume $t=0$, substitute eqn.~\eqref{eq:conditional_velocity} into eqn.~\eqref{eq:velocity_mixture}  and choose one of the common noise schedulers s.t. $\dot{\alpha}_0=1$ and $\dot{\beta}_0=-1$:
\begin{align}
  v(y_0, 0 \mid x) &= \sum_{k=1}^{K_x} w_k^x\, (\dot{\alpha}_0\, \hat{y}_k^x + \dot{\beta}_0\, y_0) = \sum_{k=1}^{K_x} w_k^x\, (\hat{y}_k^x - y_0)
\label{eq:velocity_simple}
\end{align}

\begin{figure}
    \centering
    \includegraphics[width=\textwidth]{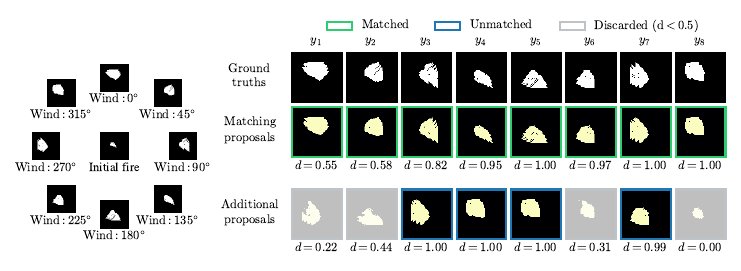}
    \caption{\textbf{MMFire proposals:} MMFire contains multiple simulated wildfire spread outcomes, differing by wind direction. Our mode proposal network successfully produces proposals covering all modes. This model was trained in the single-label scenario, only seeing a single label $y_i$ per input at each training step, sampled according to a highly-skewed distribution with probabilities ranging from $0.4\%$ for 0° to $50.2\%$ for 315°.}
    \label{fig:mmfire-proposals}
\end{figure}

\paragraph{Estimating mode probabilities via constrained linear least squares}
We assume we have access to a neural network $v_{\theta}(y_t, t \mid x)$ accurately estimating the velocity field. Then eqn.~\eqref{eq:velocity_simple} imposes this constraint on $\{w_k\}$:
\begin{equation}
v_\theta(y_0, 0 \mid x)
\approx
\sum_{k=1}^{K_x} w_k^x\, (\hat{y}_k^x - y_0)
\label{eq:velocity_approx}
\end{equation}
We can estimate $\{w_k^x\}$ by solving the following constrained linear least squares problem. Given $n$ noise samples $y^{(i)}_0 \sim N(0, I)$ then solve this constrained optimization problem.
\begin{align}
\underset{\{w_k^x\}}{\arg\min}
\;
\sum_{i=1}^n
\left\|
v_\theta(y_0^{(i)}, 0 \mid x)
-
\sum_{k=1}^{K_x} w_k^x \, (\hat{y}_k^x - y_0^{(i)})
\right\|^2
\;
\text{s.t. } w_k^x \ge 0,\ \sum_{k=1}^{K_x} w_k^x = 1.
\label{eq:pi_optimization}
\end{align}

\section{Experiments}
\label{sec:experiments}


\subsection{Datasets}

We evaluate on MMFire~\cite{gerard_wildfire_2025} (wildfire spread prediction), binary ambiguous City\-scapes~\cite{gerard_wildfire_2025}, and LIDC~\cite{armato_iii_lung_2011} (medical imaging). Each dataset has multiple segmentation masks per input. Each mask has a prior probability. These distributions are highly skewed for MMFire and Cityscapes, imitating real-life datasets where similar inputs yield diverse outcomes of varying frequencies.

In the single-label scenario  masks are sampled according to these probabilities. For duplicate masks, probability mass is summed (\eg two identical annotations in LIDC are seen as a single mask with probability $25\%+25\% = 50\%$). These summed probabilities also serve as targets during probability estimation. 

\begin{figure}
    \centering
    \includegraphics[]{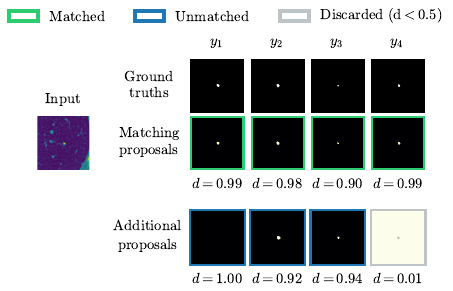}
    \caption{\textbf{LIDC proposals:} LIDC consists of expert annotations of lung CT images. Ground truths are usually white round shapes in the center of the image, making it very difficult for the model to discard implausible proposals. This model was trained in the single-label scenario. The bottom-right proposal likely did not receive many gradient updates, resulting in a bad value range, making it easy to discard.
    }
    \label{fig:lidc-proposals}
\end{figure}

\subsection{Experimental Setup}
\label{sec:experimental-setup}

We use the NSCN++ U-Net~\cite{song2021scorebased}, diffusion-pretrained in WSS~\cite{gerard_wildfire_2025} (made available upon request by the authors), both as the starting point for the model proposal model, only replacing the final convolutional layer, and as the flow model for velocity decomposition. The selection head $h_{\psi_1}$ has a DCGAN~\cite{radford_unsupervised_2015} discriminator architecture and is trained from scratch. When computing the scores $d_j$, all proposals $\{\hat{y}_{i}^x\}_i$ are considered at once, allowing for the detection of duplicates. See the supplementary material for further experimental details.

Ablations use MMFire. The main results (\cref{tab:main-results,tab:main-results-lidc}) use the test sets, all others use the validation sets. All metrics are the mean over five runs. 

To evaluate mode coverage, we use the HM IoU*~\cite{gerard_wildfire_2025} metric, which is the Hungarian-matched IoU (HM IoU) without duplicate ground truth masks. We only measure mode coverage, not the ability to correctly duplicate masks. 

When estimating probability distributions via velocity decomposition, we evaluate the results via the \textit{Brier Skill Score} (BSS):

\begin{equation}
\label{eq:bss}
    \mathrm{BSS} = 1 - \frac{\mathrm{BS}(\boldsymbol{\hat{w}}, \boldsymbol{w}_{\mathrm{true}})}{\mathrm{BS}(\boldsymbol{w}_{\mathrm{uniform}}, \boldsymbol{w}_{\mathrm{true}}) + \epsilon} \in [-\infty,1], 
\end{equation}

where the Brier Score $\mathrm{BS}(\boldsymbol{\hat{w}}, \boldsymbol{w}_{\mathrm{true}})$ is the mean squared error between estimated and ground truth probabilities, $\mathrm{BS}(\boldsymbol{w}_{\mathrm{uniform}}, \boldsymbol{w}_{\mathrm{true}})$ is obtained by predicting the uniform distribution, and $\epsilon=10^{-8}$ prevents division by zero.

\begin{table}[t]
\centering
\caption{\textbf{Main results across datasets:} Runtimes should be taken as a rough measure, since they represent the total runtime, measured on shared infrastructure.}
\label{tab:main-results}
\begin{tabular}{llcccc}
\toprule
Dataset & Method & Proposals & HM IoU*~~& Runtime \\
\midrule
MMFire & ProbU-Net~\cite{gerard_wildfire_2025} & 16 & 0.637 & 0h12m \\
       & ProbU-Net~\cite{gerard_wildfire_2025} & 32 & 0.687 & 0h27m \\
       & Diffusion + SPELL~\cite{gerard_wildfire_2025} & 16 & 0.784 & 0h55m\\
       & Diffusion + SPELL~\cite{gerard_wildfire_2025} & 32 & 0.830 & 2h17m \\
       \cmidrule{2-5}
       & ModeProposalNet (Single) & 16 & 0.909 & 0h2m \\
       & ModeProposalNet (Full) & 16 & \textbf{0.923} & 0h2m \\
\midrule
Cityscapes~ & ProbU-Net~\cite{gerard_wildfire_2025}  & 32 & 0.427 & 0h12m \\
            & ProbU-Net~\cite{gerard_wildfire_2025}  & 64 & 0.479 & 0h27m \\
            & Diffusion + PG~\cite{gerard_wildfire_2025} & 32 & 0.696 & 1h37m  \\
            & Diffusion + PG~\cite{gerard_wildfire_2025} & 64 & 0.738 & 3h40m \\
            \cmidrule{2-5}
            & ModeProposalNet (Single) & 32 & 0.813 & 0h3m\\
            & ModeProposalNet (Full) & 32 & \textbf{0.832} & 0h3m\\
\midrule
LIDC & ProbU-Net~\cite{gerard_wildfire_2025} & 8 & 0.715 & 0h7m \\
     & ProbU-Net~\cite{gerard_wildfire_2025} & 16 & 0.785 & 0h13m \\
     & Diffusion + SPELL~\cite{gerard_wildfire_2025} & 8 & 0.697 & 1h41m\\
     & Diffusion + SPELL~\cite{gerard_wildfire_2025} & 16 & 0.751 & 3h31m \\
     \cmidrule{2-5}
     & ModeProposalNet (Single) & 8 & 0.814 & 0h5m \\
     & ModeProposalNet (Full) & 8 & \textbf{0.818} & 0h5m \\
\bottomrule

\end{tabular}
\end{table}

\subsection{Learning Strong Proposals}
\label{sec:res-learn}

\textbf{Fully-labeled training} The mode proposal model achieves a high performance of 92.7\% HM IoU* on MMFire. Retaining features from diffusion pre-training, by freezing the encoder, increases the selection F1 score by 4.2\% (\cref{tab:freeze-weights}, suppl.), compared to training all parameters, with little change in HM IoU*. Consequently, we freeze the encoder and only train the decoder. 

\textbf{Single-label training} The performance stays within $2\%$ HM IoU* of the fully-labeled model (\cref{tab:main-results}), even though each training step only adjusts a single proposal per batch item. Example proposals are shown in \cref{fig:mmfire-proposals} (MMFire), \cref{fig:lidc-proposals} (LIDC) and \cref{fig:cs-proposals} (Cityscapes, suppl.).

\textbf{Comparison to WSS} Mode proposal models greatly increase the HM IoU* compared to diversity-guided diffusion (\cref{tab:main-results}). While ProbUNet may match our performance on LIDC by more exhaustive sampling, MMFire's highly skewed distribution would require $>300$ samples to cover all modes. To generate 32 samples for each point in the test set, ProbUNet takes tens of minutes, while the diffusion samplers take hours due to the iterative denoising. In contrast, our models reach a higher mode coverage within minutes. 

\textbf{Comparison to LIDC literature} The default metric for LIDC is HM IoU, measuring whether ground truth masks are reproduced accurately and according to their relative frequency. This creates a mismatch with our approach, which is explicitly trained to produce unique modes instead of duplicates. Despite this, \cref{tab:main-results-lidc} shows that our model with eight proposals and single-label training outperforms most competing generative methods. 
By removing the selection loss in training ($\lambda=0$) and increasing the proposal count, we achieve performance within 0.8\% of the best competing method, 
using only a single deterministic forward pass. While increasing the number of proposals greatly boosts our HM IoU score, the $\hmioum$ score reveals that this essentially only adds duplicates, confirming the ability of our method to learn the primary modes of a distribution even with a low number of proposals.

\begin{table}[t]
\centering
\caption{\textbf{Comparison to LIDC literature:} We compare performances on the LIDC test set. HM IoU is the standard metric, \ie without deduplication of ground truths. In $\hmioum$, duplicate ground truths are matched to the same proposal, removing the need to produce duplicate proposals. This is a more appropriate measure for our approach, since our training does not encourage duplicate proposals in any way.}
\label{tab:main-results-lidc}
\begin{tabular}{lccc}
\toprule
Method & Proposals ~~& HM IoU~~& $\hmioum$\\
\midrule
CIMD\cite{rahman_ambiguous_2023}, reported in \cite{zbinden_stochastic_2023} & 32 & 0.592 \\
BerDiff~\cite{chen_berdiff_2023} & 16 & 0.596 \\
CCDM~\cite{zbinden_stochastic_2023} & 32 & 0.631 \\
MoSE\cite{gao2023modeling} & 100 & 0.638 \\
JProbU-Net~\cite{zhang_probabilistic_2022} & 32 & 0.647\\
P2SAM~\cite{huang_p2sam_2024} & 16 & 0.679 \\
FlowSSN~\cite{de2025flow} & 16 & \textbf{0.879}\\
\midrule
ModeProposalNet (Single) & 8 & 0.652 & 0.846 \\
ModeProposalNet (Full) & 8 & 0.654  & 0.848 \\
ModeProposalNet (Full, $\lambda=0$) & 16 & 0.712 & 0.855 \\
ModeProposalNet (Full, $\lambda=0$) & 32 & 0.810 & 0.866 \\
ModeProposalNet (Full, $\lambda=0$) & 64 & \textbf{0.871} & \textbf{0.875} \\
\bottomrule

\end{tabular}
\end{table}

\subsection{Selecting the Best Proposals}
\label{sec:res-selection}

While the HM IoU* metric is unimpeded by superfluous proposals, any real-life application would likely benefit from reducing the number of proposals that need to processed for downstream tasks. To achieve this, we have adopted the mechanism of per-proposal selection scores from DETR~\cite{carion2020end}.

In the \textbf{fully-labeled} scenario, all models have learned to discard the correct number of proposals via selection scores (\cref{tab:filter-proposals-all-datasets}). Since Hungarian matching always picks the best-matching proposals from those available, discarding proposals can only reduce the segmentation performance. For MMFire, we only lose $2.3\%$ HM IoU*, for Cityscapes $2.6\%$. This strong result confirms the practical utility of the selection scores, given that half of the proposals are discarded at inference time, solely based on the learned scores. 

\begin{table}[]
    \centering
    \caption{{MMFire: Filtering superfluous proposals by selection scores.}}
    \begin{tabular}{lcccc}
    \toprule
Scenario~~ &  Filter duplicates & Selection F1  & HM IoU* ~~& Proposals \\\midrule
\addlinespace
& & \multicolumn{2}{c}{MMFire - 8 modes} &\\\midrule
Full & \xmark & \textbf{0.994} & \textbf{0.923} & 16.0\\
Full & \cmark & & 0.920 & \textbf{7.9} \\\cmidrule{2-5}
Single & \xmark & 0.692 & 0.910 & 16.0 \\
Single & \cmark &  & 0.894 & 14.0 \\ 
\addlinespace
& & \multicolumn{2}{c}{Cityscapes - 16 modes} &\\\midrule
Full & \xmark & \textbf{0.901} & \textbf{0.842} & 32.0 \\
Full & \cmark & & 0.816 & \textbf{15.8}\\\cmidrule{2-5}
Single & \xmark & 0.760 & 0.820 & 32.0 \\
Single & \cmark &  & 0.813 & 21.4 \\
\addlinespace
& & \multicolumn{2}{c}{LIDC - 4 modes} &\\\midrule
Full & \xmark & 0.504 & \textbf{0.812} & 8.0 \\
Full & \cmark & & 0.582 & \textbf{4.1}\\\cmidrule{2-5}
Single & \xmark & 0.579 & 0.807 & 8.0 \\
Single & \cmark & & 0.805 & 7.8 \\ 
\bottomrule
\end{tabular}
    \label{tab:filter-proposals-all-datasets}
\end{table}

In the \textbf{single-labeled} scenario, the MMFire model only discards two proposals on average, at a cost of $1.6\%$ HM IoU*, while the Cityscapes model still discards $8.6$, costing $0.7\%$. For Cityscapes, some proposal heads are under-trained, likely not being matched often enough during training. They are correctly discarded by the selection scores. \cref{fig:cs-proposals} (suppl.) shows an example batch. 

These results demonstrate that the PU loss allows models to learn sensible scores in this very weakly-labeled scenario. However, near-duplicates are not reliably discarded anymore (\eg \cref{fig:mmfire-proposals}). In the fully-labeled scenario, they are easily detected because Hungarian matching with all ground truths provides a selection label for \textit{every} proposal. However, in the single-label scenario, we only have a \textit{single} positive selection label, while all others are treated as soft negatives by the PU loss, which does not provide any de-duplication signal.

\textbf{Special case: LIDC} Selection scores perform badly on LIDC, with F1 scores of 50.4\% (full) and 57.9\% (single), likely because of low intra-modal variance. Modes in MMFire and Cityscapes differ strongly, making it easy to detect both ill-fitting and duplicate proposals. In contrast, LIDC's modes tend to be circular shapes in the image center, varying only slightly in their extent, which makes this distinction inherently difficult. \cref{fig:lidc-proposals} visualizes this.

\begin{table}[]
\begin{minipage}[t]{.45\textwidth}
    \centering
    \caption{\textbf{Ablation: Velocity decomposition} into sum over ground truths instead of proposals.}
    \begin{tabular}{lcc}
    \toprule
    Dataset & BSS $\uparrow$ & Runtime\\\midrule
    MMFire & 0.919 & 10m\\
    CS (underconstrained) & 0.443 & 09m\\
    CS (structure-aware) & 0.870 & 14m\\
    LIDC &  <0 & 33m\\\bottomrule
    \end{tabular}
    \label{tab:velocity-decomp-gt-masks}
    \end{minipage}\hfill
    \begin{minipage}[t]{.53\textwidth}
    \centering
    \caption{\textbf{Filter duplicate MMFire proposals} to remove undesired solutions from velocity decomposition.}
    \begin{tabular}{lccc}
    \toprule
    Scenario ~& Filter scores ~& Filter dup. ~& BSS $\uparrow$\\\midrule
    Full & \xmark & \xmark & 0.924 \\
         & \cmark & \xmark & \textbf{0.940} \\
         & \cmark & \cmark & 0.939 \\
    \midrule
    Single & \xmark & \xmark & 0.205 \\
         & \cmark & \xmark & 0.199 \\
         & \cmark & \cmark & \textbf{0.821} \\\midrule
    \multicolumn{2}{c}{Ground truths}  & & \textbf{0.919} \\\bottomrule
    \end{tabular}
    \label{tab:velocity-decomp-filter-proposals}
\end{minipage}
\end{table}

\subsection{Estimating Prior Mode Probabilities}
\label{sec:res-estimate-probabilities}

After producing a filtered-down set of proposals, we now try to additionally estimate the prior probability of each proposal. This way, we would be able to replace generative models for ambiguous segmentation, but at a fraction of the inference cost. See \cref{sec:method-estimate-probabilities} for method details.

\textbf{Upper bound using ground truth} We decompose the learned velocity into a weighted sum using the \textit{ground truth} modes of the distribution, instead of imperfect proposals. \cref{tab:velocity-decomp-gt-masks} shows that the probabilities estimated for MMFire are close to correct ($91.9\%$ BSS). For Cityscapes, the specific structure underlying the modes leads to the least squares problem being underconstrained, resulting in a BSS of $44.3\%$. Integrating knowledge about this underlying structure (\cref{sec:supp-cs-estimate-probabilities},  suppl.) resolves this issue ($87.0\%$ BSS). These results demonstrate that velocity decomposition can work well in principle. 

For LIDC, our approach is inferior to the uniform baseline. The reasons are two-fold: 1. LIDC's segmentation masks mostly consist of round centered shapes, making it difficult to decompose the velocity into a sum over distinct shapes. 2. For LIDC-images without duplicate ground truth masks, the uniform distribution \textit{is} the ground truth distribution.

\textbf{Fully-labeled scenario} Models trained in the fully-labeled scenario achieve a high BSS of $92.4\%$, even without filtering proposals (\cref{tab:velocity-decomp-filter-proposals}). Using the selection scores adds $1.6\%$, reaffirming their utility. Manual inspection of the success without filtering shows that many of the proposals are very low quality, likely because the training quickly focused on updating a subset of proposals, leaving the others untrained and thus not useful for the least squares problem.

For Cityscapes, we demonstrate in \cref{sec:supp-cs-estimate-probabilities} (suppl.) that the structure-aware approach also produces strong estimates from mode proposals (single- or fully-labeled). However, since this approach requires a manual thresholding step for each set of proposals, we only show it exemplarily for one set of proposals.

\textbf{Single-label scenario} Probability estimates perform badly (BSS of $\leq20.5\%$), both for the full set of proposals, and the subset filtered by selection scores, consistent with the low selection F1 of $69.2\%$ (\cref{tab:filter-proposals-all-datasets}). However, removing near-duplicates (IoU $\geq 0.95$) increases the BSS by $+61.6\%$. Manual inspection reveals that the numerical solver spreads mass across similar proposals. De-duplication removes this ambiguity. These results on MMFire confirm that, for sufficiently distinct modes, velocity decomposition can approximate prior probabilities from imperfect proposals, without the high sampling cost of generative models.

\section{Discussion and Limitations}

\textbf{Limitation of velocity decomposition} While velocity decomposition succeeds on MMFire and Cityscapes, it fails on LIDC, suggesting that the approach is limited to datasets where modes exhibit sufficient diversity. We observed that the solver distributes the probability mass across similar-looking proposals, instead of deciding on one, which supports the conclusion that this method requires distinct masks to work well. Furthermore, we found that the structure underlying Cityscapes' modes led to the least squares problem being underdetermined, resulting in undesired solutions. However, we showed that if the underlying structure is known, it can potentially be integrated, allowing velocity decomposition to resolve the underdetermination and find the desired prior distribution. 

\textbf{Datasets} Our experiments are limited by the available datasets. Ambiguous segmentation is mostly discussed in the medical imaging literature, where dataset characteristics like intra-mode diversity are similar to LIDC. However, the fundamental problem of aleatoric uncertainty in segmentation occurs outside of medical imaging as well, \eg in future predictions related to Earth observation: wildfire spread, land slides, or flooding. The number of public datasets addressing the ambiguity aspect of these tasks is unfortunately very limited.

\textbf{Utility of generative models} Our modeling approach implicitly assumes that the ambiguity in the predictions can essentially be characterized as a categorical distribution over a low number of discrete modes. This assumption fails when the output distribution is more continuous, such that a low fixed number of proposals can not capture the underlying diversity appropriately. Furthermore, if the generation itself is challenging, iterative sampling approaches employed by diffusion or flow matching models may produce higher quality predictions. 


\section{Conclusion}

In this work, we introduced mode proposal models, an efficient alternative to generative models for ambiguous segmentation. We demonstrated that proposal heads, trained with Hungarian-matched loss, can often achieve a better mode coverage than generative models, using fewer samples and less compute. Furthermore, we showed that these diverse proposals can even be learned in the single-label scenario, which is common in real-world observational data. To discard redundant proposals, we adopted a selection mechanism from object detection and demonstrated that it can discard up to half of the proposals with only a small impact on mode coverage. Finally, we introduced and validated velocity decomposition as an approach to estimate prior mode probabilities without high-volume sampling, assuming the underlying modes are sufficiently distinct. Future work will focus on applying these advances to observational real-world datasets and evaluating their impact.

\section*{Acknowledgements}
This work is funded by Digital Futures in the project EO-AI4GlobalChange. The computations were enabled by resources provided by the National Academic Infrastructure for Supercomputing in Sweden (NAISS) at C3SE partially funded by the Swedish Research Council through grant agreement no. 2022-06725.

%
%
\bibliographystyle{splncs04}
\bibliography{main}

@String(CVPR  = {IEEE Conf. Comput. Vis. Pattern Recog.})

@String(ICCV  = {Int. Conf. Comput. Vis.})

@String(CVPR  = {CVPR})

@String(ICCV  = {ICCV})

@inproceedings{gerard_wildfire_2025,
    title = {Wildfire {Spread} {Scenarios}: {Increasing} {Sample} {Diversity} of {Segmentation} {Diffusion} {Models} with {Training}-{Free} {Methods}},
    shorttitle = {Wildfire {Spread} {Scenarios}},
    url = {https://openreview.net/forum?id=E44d5hzEV0#discussion},
    language = {en},
    urldate = {2025-11-09},
    author = {Gerard, Sebastian and Sullivan, Josephine},
    month = nov,
    year = {2025},
}

@inproceedings{kiryo_positive-unlabeled_2017,
    address = {Red Hook, NY, USA},
    series = {{NIPS}'17},
    title = {Positive-unlabeled learning with non-negative risk estimator},
    isbn = {978-1-5108-6096-4},
    url = {https://dl.acm.org/doi/10.5555/3294771.3294931},
    urldate = {2026-02-15},
    booktitle = {Proceedings of the 31st {International} {Conference} on {Neural} {Information} {Processing} {Systems}},
    publisher = {Curran Associates Inc.},
    author = {Kiryo, Ryuichi and Niu, Gang and du Plessis, Marthinus C. and Sugiyama, Masashi},
    month = dec,
    year = {2017},
    pages = {1674--1684},
}

@article{radford_unsupervised_2015,
    title = {Unsupervised {Representation} {Learning} with {Deep} {Convolutional} {Generative} {Adversarial} {Networks}},
    url = {https://www.semanticscholar.org/paper/Unsupervised-Representation-Learning-with-Deep-Radford-Metz/8388f1be26329fa45e5807e968a641ce170ea078},
    urldate = {2026-02-15},
    journal = {CoRR},
    author = {Radford, Alec and Metz, Luke and Chintala, Soumith},
    month = nov,
    year = {2015},
}

@inproceedings{zbinden_stochastic_2023,
    address = {Paris, France},
    title = {Stochastic {Segmentation} with {Conditional} {Categorical} {Diffusion} {Models}},
    copyright = {https://doi.org/10.15223/policy-029},
    isbn = {979-8-3503-0718-4},
    url = {https://ieeexplore.ieee.org/document/10376866/},
    doi = {10.1109/ICCV51070.2023.00109},
    language = {en},
    urldate = {2025-01-27},
    booktitle = {2023 {IEEE}/{CVF} {International} {Conference} on {Computer} {Vision} ({ICCV})},
    publisher = {IEEE},
    author = {Zbinden, Lukas and Doorenbos, Lars and Pissas, Theodoros and Huber, Adrian Thomas and Sznitman, Raphael and Márquez-Neila, Pablo},
    month = oct,
    year = {2023},
    pages = {1119--1129},
}

@inproceedings{gao2023modeling,
    title = {Modeling multimodal aleatoric uncertainty in segmentation with mixture of stochastic experts},
    url = {https://openreview.net/forum?id=KE_wJD2RK4},
    booktitle = {The eleventh international conference on learning representations},
    author = {Gao, Zhitong and Chen, Yucong and Zhang, Chuyu and He, Xuming},
    year = {2023},
}

@inproceedings{carion2020end,
    title = {End-to-end object detection with transformers},
    booktitle = {European conference on computer vision},
    publisher = {Springer},
    author = {Carion, Nicolas and Massa, Francisco and Synnaeve, Gabriel and Usunier, Nicolas and Kirillov, Alexander and Zagoruyko, Sergey},
    year = {2020},
    pages = {213--229},
}

@inproceedings{zhang_probabilistic_2022,
    address = {New York, NY, USA},
    series = {{MM} '22},
    title = {A {Probabilistic} {Model} for {Controlling} {Diversity} and {Accuracy} of {Ambiguous} {Medical} {Image} {Segmentation}},
    isbn = {978-1-4503-9203-7},
    url = {https://dl.acm.org/doi/10.1145/3503161.3548115},
    doi = {10.1145/3503161.3548115},
    urldate = {2026-02-26},
    booktitle = {Proceedings of the 30th {ACM} {International} {Conference} on {Multimedia}},
    publisher = {Association for Computing Machinery},
    author = {Zhang, Wei and Zhang, Xiaohong and Huang, Sheng and Lu, Yuting and Wang, Kun},
    month = oct,
    year = {2022},
    pages = {4751--4759},
}

@inproceedings{cordts_cityscapes_2016,
    title = {The {Cityscapes} {Dataset} for {Semantic} {Urban} {Scene} {Understanding}},
    url = {https://www.cv-foundation.org/openaccess/content_cvpr_2016/html/Cordts_The_Cityscapes_Dataset_CVPR_2016_paper.html},
    urldate = {2025-03-02},
    author = {Cordts, Marius and Omran, Mohamed and Ramos, Sebastian and Rehfeld, Timo and Enzweiler, Markus and Benenson, Rodrigo and Franke, Uwe and Roth, Stefan and Schiele, Bernt},
    year = {2016},
    pages = {3213--3223},
}

@article{armato_iii_lung_2011,
    title = {The {Lung} {Image} {Database} {Consortium} ({LIDC}) and {Image} {Database} {Resource} {Initiative} ({IDRI}): {A} {Completed} {Reference} {Database} of {Lung} {Nodules} on {CT} {Scans}},
    volume = {38},
    copyright = {© 2011 American Association of Physicists in Medicine},
    issn = {2473-4209},
    shorttitle = {The {Lung} {Image} {Database} {Consortium} ({LIDC}) and {Image} {Database} {Resource} {Initiative} ({IDRI})},
    url = {https://onlinelibrary.wiley.com/doi/abs/10.1118/1.3528204},
    doi = {10.1118/1.3528204},
    language = {en},
    number = {2},
    urldate = {2025-02-26},
    journal = {Medical Physics},
    author = {Armato III, Samuel G. and McLennan, Geoffrey and Bidaut, Luc and McNitt-Gray, Michael F. and Meyer, Charles R. and Reeves, Anthony P. and Zhao, Binsheng and Aberle, Denise R. and Henschke, Claudia I. and Hoffman, Eric A. and Kazerooni, Ella A. and MacMahon, Heber and van Beek, Edwin J. R. and Yankelevitz, David and Biancardi, Alberto M. and Bland, Peyton H. and Brown, Matthew S. and Engelmann, Roger M. and Laderach, Gary E. and Max, Daniel and Pais, Richard C. and Qing, David P.-Y. and Roberts, Rachael Y. and Smith, Amanda R. and Starkey, Adam and Batra, Poonam and Caligiuri, Philip and Farooqi, Ali and Gladish, Gregory W. and Jude, C. Matilda and Munden, Reginald F. and Petkovska, Iva and Quint, Leslie E. and Schwartz, Lawrence H. and Sundaram, Baskaran and Dodd, Lori E. and Fenimore, Charles and Gur, David and Petrick, Nicholas and Freymann, John and Kirby, Justin and Hughes, Brian and Vande Casteele, Alessi and Gupte, Sangeeta and Sallam, Maha and Heath, Michael D. and Kuhn, Michael H. and Dharaiya, Ekta and Burns, Richard and Fryd, David S. and Salganicoff, Marcos and Anand, Vikram and Shreter, Uri and Vastagh, Stephen and Croft, Barbara Y. and Clarke, Laurence P.},
    year = {2011},
    pages = {915--931},
}

@inproceedings{guzman-rivera_multiple_2012,
    title = {Multiple {Choice} {Learning}: {Learning} to {Produce} {Multiple} {Structured} {Outputs}},
    volume = {25},
    shorttitle = {Multiple {Choice} {Learning}},
    url = {https://proceedings.neurips.cc/paper_files/paper/2012/hash/cfbce4c1d7c425baf21d6b6f2babe6be-Abstract.html},
    urldate = {2026-03-03},
    booktitle = {Advances in {Neural} {Information} {Processing} {Systems}},
    publisher = {Curran Associates, Inc.},
    author = {Guzmán-rivera, Abner and Batra, Dhruv and Kohli, Pushmeet},
    year = {2012},
}

@inproceedings{lee_stochastic_2016,
    title = {Stochastic {Multiple} {Choice} {Learning} for {Training} {Diverse} {Deep} {Ensembles}},
    volume = {29},
    url = {https://proceedings.neurips.cc/paper_files/paper/2016/hash/20d135f0f28185b84a4cf7aa51f29500-Abstract.html},
    urldate = {2026-03-03},
    booktitle = {Advances in {Neural} {Information} {Processing} {Systems}},
    publisher = {Curran Associates, Inc.},
    author = {Lee, Stefan and Purushwalkam Shiva Prakash, Senthil and Cogswell, Michael and Ranjan, Viresh and Crandall, David and Batra, Dhruv},
    year = {2016},
}

@inproceedings{tian_versatile_2019,
    address = {Long Beach, CA, USA},
    title = {Versatile {Multiple} {Choice} {Learning} and {Its} {Application} to {Vision} {Computing}},
    copyright = {https://doi.org/10.15223/policy-029},
    isbn = {978-1-7281-3293-8},
    url = {https://ieeexplore.ieee.org/document/8953986/},
    doi = {10.1109/CVPR.2019.00651},
    language = {en},
    urldate = {2026-03-03},
    booktitle = {2019 {IEEE}/{CVF} {Conference} on {Computer} {Vision} and {Pattern} {Recognition} ({CVPR})},
    publisher = {IEEE},
    author = {Tian, Kai and Xu, Yi and Zhou, Shuigeng and Guan, Jihong},
    month = jun,
    year = {2019},
    pages = {6342--6350},
}

@inproceedings{kohl_probabilistic_2018,
    title = {A {Probabilistic} {U}-{Net} for {Segmentation} of {Ambiguous} {Images}},
    volume = {31},
    url = {https://proceedings.neurips.cc/paper/2018/hash/473447ac58e1cd7e96172575f48dca3b-Abstract.html},
    urldate = {2024-11-12},
    booktitle = {Advances in {Neural} {Information} {Processing} {Systems}},
    publisher = {Curran Associates, Inc.},
    author = {Kohl, Simon and Romera-Paredes, Bernardino and Meyer, Clemens and De Fauw, Jeffrey and Ledsam, Joseph R. and Maier-Hein, Klaus and Eslami, S. M. Ali and Jimenez Rezende, Danilo and Ronneberger, Olaf},
    year = {2018},
}

@inproceedings{rahman_ambiguous_2023,
    title = {Ambiguous {Medical} {Image} {Segmentation} {Using} {Diffusion} {Models}},
    booktitle = {Proceedings of the {IEEE}/{CVF} {Conference} on {Computer} {Vision} and {Pattern} {Recognition} ({CVPR})},
    author = {Rahman, Aimon and Valanarasu, Jeya Maria Jose and Hacihaliloglu, Ilker and Patel, Vishal M.},
    month = jun,
    year = {2023},
    pages = {11536--11546},
}

@inproceedings{monteiro_stochastic_2020,
    title = {Stochastic {Segmentation} {Networks}: {Modelling} {Spatially} {Correlated} {Aleatoric} {Uncertainty}},
    volume = {33},
    shorttitle = {Stochastic {Segmentation} {Networks}},
    url = {https://proceedings.neurips.cc/paper/2020/hash/95f8d9901ca8878e291552f001f67692-Abstract.html},
    urldate = {2026-03-03},
    booktitle = {Advances in {Neural} {Information} {Processing} {Systems}},
    publisher = {Curran Associates, Inc.},
    author = {Monteiro, Miguel and Le Folgoc, Loic and Coelho de Castro, Daniel and Pawlowski, Nick and Marques, Bernardo and Kamnitsas, Konstantinos and van der Wilk, Mark and Glocker, Ben},
    year = {2020},
    pages = {12756--12767},
}

@misc{lipman_flow_2024,
    title = {Flow {Matching} {Guide} and {Code}},
    url = {http://arxiv.org/abs/2412.06264},
    doi = {10.48550/arXiv.2412.06264},
    urldate = {2026-03-04},
    publisher = {arXiv},
    author = {Lipman, Yaron and Havasi, Marton and Holderrieth, Peter and Shaul, Neta and Le, Matt and Karrer, Brian and Chen, Ricky T. Q. and Lopez-Paz, David and Ben-Hamu, Heli and Gat, Itai},
    month = dec,
    year = {2024},
    note = {arXiv:2412.06264 [cs]},
}

@inproceedings{schusterbauer_diff2flow_2025,
    title = {{Diff2Flow}: {Training} {Flow} {Matching} {Models} via {Diffusion} {Model} {Alignment}},
    shorttitle = {{Diff2Flow}},
    url = {https://ieeexplore.ieee.org/document/11092660},
    doi = {10.1109/CVPR52734.2025.02640},
    urldate = {2026-03-04},
    booktitle = {2025 {IEEE}/{CVF} {Conference} on {Computer} {Vision} and {Pattern} {Recognition} ({CVPR})},
    author = {Schusterbauer, Johannes and Gui, Ming and Fundel, Frank and Ommer, Björn},
    month = jun,
    year = {2025},
    note = {ISSN: 2575-7075},
    pages = {28347--28357},
}

@inproceedings{song2021scorebased,
    title = {Score-based generative modeling through stochastic differential equations},
    url = {https://openreview.net/forum?id=PxTIG12RRHS},
    booktitle = {International conference on learning representations},
    author = {Song, Yang and Sohl-Dickstein, Jascha and Kingma, Diederik P and Kumar, Abhishek and Ermon, Stefano and Poole, Ben},
    year = {2021},
}

@inproceedings{chen_berdiff_2023,
    title = {Berdiff: {Conditional} bernoulli diffusion model for medical image segmentation},
    booktitle = {International {Conference} on {Medical} {Image} {Computing} and {Computer}-{Assisted} {Intervention}},
    publisher = {Springer},
    author = {Chen, Tao and Wang, Chenhui and Shan, Hongming},
    year = {2023},
    pages = {491--501},
}

@inproceedings{huang_p2sam_2024,
    address = {New York, NY, USA},
    series = {{MM} '24},
    title = {{P2SAM}: {Probabilistically} {Prompted} {SAMs} {Are} {Efficient} {Segmentator} for {Ambiguous} {Medical} {Images}},
    isbn = {979-8-4007-0686-8},
    shorttitle = {{P2SAM}},
    url = {https://dl.acm.org/doi/10.1145/3664647.3680628},
    doi = {10.1145/3664647.3680628},
    urldate = {2026-03-04},
    booktitle = {Proceedings of the 32nd {ACM} {International} {Conference} on {Multimedia}},
    publisher = {Association for Computing Machinery},
    author = {Huang, Yuzhi and Li, Chenxin and Lin, Zixu and Liu, Hengyu and Xu, Haote and Liu, Yifan and Huang, Yue and Ding, Xinghao and Tu, Xiaotong and Yuan, Yixuan},
    month = oct,
    year = {2024},
    pages = {9779--9788},
}

@inproceedings{de2025flow,
    title = {Flow stochastic segmentation networks},
    booktitle = {Proceedings of the {IEEE}/{CVF} international conference on computer vision},
    author = {De Sousa Ribeiro, Fabio and Todd, Omar and Jones, Charles and Kori, Avinash and Mehta, Raghav and Glocker, Ben},
    year = {2025},
    pages = {14754--14765},
}

@inproceedings{kirillov_segment_2023,
    address = {Paris, France},
    title = {Segment {Anything}},
    copyright = {https://doi.org/10.15223/policy-029},
    isbn = {979-8-3503-0718-4},
    url = {https://ieeexplore.ieee.org/document/10378323/},
    doi = {10.1109/ICCV51070.2023.00371},
    language = {en},
    urldate = {2026-03-04},
    booktitle = {2023 {IEEE}/{CVF} {International} {Conference} on {Computer} {Vision} ({ICCV})},
    publisher = {IEEE},
    author = {Kirillov, Alexander and Mintun, Eric and Ravi, Nikhila and Mao, Hanzi and Rolland, Chloe and Gustafson, Laura and Xiao, Tete and Whitehead, Spencer and Berg, Alexander C. and Lo, Wan-Yen and Dollár, Piotr and Girshick, Ross},
    month = oct,
    year = {2023},
    pages = {3992--4003},
}

@inproceedings{corso_particle_2023,
    title = {Particle {Guidance}: non-{I}.{I}.{D}. {Diverse} {Sampling} with {Diffusion} {Models}},
    shorttitle = {Particle {Guidance}},
    url = {https://openreview.net/forum?id=hEyIHsyZ9F},
    language = {en},
    urldate = {2024-11-24},
    author = {Corso, Gabriele and Xu, Yilun and Bortoli, Valentin De and Barzilay, Regina and Jaakkola, Tommi},
    month = nov,
    year = {2023},
}

@inproceedings{kirchhof_shielded_2025,
    title = {Shielded {Diffusion}: {Generating} {Novel} and {Diverse} {Images} using {Sparse} {Repellency}},
    shorttitle = {Shielded {Diffusion}},
    url = {https://openreview.net/forum?id=XAckVo0iNj},
    language = {en},
    urldate = {2025-08-28},
    author = {Kirchhof, Michael and Thornton, James and Béthune, Louis and Ablin, Pierre and Ndiaye, Eugene and Cuturi, Marco},
    month = jun,
    year = {2025},
}

@inproceedings{de_comite_positive_1999,
    address = {Berlin, Heidelberg},
    title = {Positive and {Unlabeled} {Examples} {Help} {Learning}},
    isbn = {978-3-540-46769-4},
    doi = {10.1007/3-540-46769-6_18},
    language = {en},
    booktitle = {Algorithmic {Learning} {Theory}},
    publisher = {Springer},
    author = {De Comité, Francesco and Denis, François and Gilleron, Rémi and Letouzey, Fabien},
    editor = {Watanabe, Osamu and Yokomori, Takashi},
    year = {1999},
    pages = {219--230},
}

@inproceedings{letouzey_learning_2000,
    address = {Berlin, Heidelberg},
    title = {Learning {From} {Positive} and {Unlabeled} {Examples}},
    isbn = {978-3-540-40992-2},
    doi = {10.1007/3-540-40992-0_6},
    language = {en},
    booktitle = {Algorithmic {Learning} {Theory}},
    publisher = {Springer},
    author = {Letouzey, Fabien and Denis, François and Gilleron, Rémi},
    editor = {Arimura, Hiroki and Jain, Sanjay and Sharma, Arun},
    year = {2000},
    pages = {71--85},
}

@article{otsu_threshold_1979,
    title = {A {Threshold} {Selection} {Method} from {Gray}-{Level} {Histograms}},
    volume = {9},
    issn = {2168-2909},
    doi = {10.1109/TSMC.1979.4310076},
    number = {1},
    journal = {IEEE Transactions on Systems, Man, and Cybernetics},
    author = {Otsu, N.},
    month = jan,
    year = {1979},
    note = {Conference Name: IEEE Transactions on Systems, Man, and Cybernetics},
    pages = {62--66},
}

\newpage
\appendix

\section{Experimental Details}

We use PyTorch and PyTorch Lightning as the basis for our experiments, building on  WildfireSpreadScenarios (WSS)~\cite{gerard_wildfire_2025} and its public code. 
All reported metrics are the mean over five models, trained with different random seeds. For velocity decomposition in \cref{tab:velocity-decomp-gt-masks}, the parameters of the proposal model are not used, since these experiment use the ground truth instead of the proposals. Thus, only the underlying diffusion model is of relevance, of which we only have access to one per dataset, pretrained in ~\cite{gerard_wildfire_2025} (provided to us upon request by the authors). Therefore, these results are the mean over five runs with different random seeds, which determine at which points the velocity field is evaluated and decomposed.

\subsection{Datasets}
\label{sec:app-datasets}


\subsubsection{MMFire} is a simulated wildfire spread dataset.\cite{gerard_wildfire_2025} It consists of 9608 scenarios, comprising of seven fire-spread-related variables, including the initial state of the fire, and represented as images of shape $7\times64\times64$. For each of these initial conditions, eight outcomes were simulated, differing only by wind direction. For training and evaluating models with this dataset, each wind direction, and thereby their resulting outcomes, is assigned a prior probability. This probability distribution is highly skewed, with probabilities ranging from $0.4\%$ to $50.2\%$. While these skewed probabilities can be challenging to model, the mapping between input and output is easy to learn for this dataset, since the underlying simulator is relatively simple. Duplicate modes can occur in this dataset if terrain conditions impede wildfire spread in certain directions, \eg a lake or a fuel-free area, such that multiple wind directions lead to the same burned area. See \cref{fig:mmfire-proposals} (left) for a set of example outcomes.

\subsubsection{Cityscapes}\cite{cordts_cityscapes_2016} is a widely-used segmentation benchmark. We use the binary ambiguous segmentation variant introduced in WSS\cite{gerard_wildfire_2025}, which is down-scaled to resolution $64\times128$. To create ambiguous labels, the four classes \textit{road}, \textit{sidewalk}, \textit{vegetation}, and \textit{car} are flipped to either be part of the negative or the positive class, following per-class Bernoulli distributions with probabilities $5\%, 25\%, 75\%, 95\%$. Any pixels not belonging to these four semantic classes are always assigned to the negative class. This setup results in a highly-skewed distribution over the $2^4 = 16$ modes. Duplicate modes can occur when a semantic class is missing in an image, such that flipping its associated pixels between positive and negative has no impact. See \cref{fig:cs-proposals} for an input image and its associated set of outcomes. 

The WSS paper~\cite{gerard_wildfire_2025} performed both hyperparameter selection and evaluation on the Cityscapes validation dataset, because the test set does not have any public labels that could be transferred to the ambiguous segmentation task. We instead split off 500 random images from the training set and use this for validation and hyperparameter selection. We keep the original Cityscapes validation set intact and treat it as a test set. This way, our final results are computed on the same data as the previous works', but independent of the parameter selection. 

\subsubsection{LIDC}~\cite{armato_iii_lung_2011} consists of computed tomography lung images, each with four annotations of lung nodules, created by medical experts, displaying real-life annotation ambiguity. We follow the instructions in MoSE~\cite{gao2023modeling} to download the preprocessed dataset, sometimes referred to~\cite{zbinden_stochastic_2023} as LIDCv1, containing 9794 training images, 2314 validation images, and 2988 test images, each of size $128\times 128$ pixels. Since there is no structure within the annotations, each annotated mask is assigned a probability of $25\%$. Duplicate annotations occur often in this dataset, when annotators agreed with each other. See \cref{fig:lidc-proposals} for an example input and associated expert annotations.

\subsection{Model initialization} We initialize $f_{\phi}$ with the weights of the diffusion model trained on the respective dataset, excluding the final convolutional layer. This essentially acts as a feature extractor. The final convolutional layer that used to output a single-channel image is replaced with one that outputs $\hat{K}$ images, and becomes our set of proposal heads $g_{\psi_0}$. We initialize $\psi_0$ by duplicating the pretrained convolution weights $\hat{K}$ times, only slightly scaling each weight by a random scaling factor sampled from $U(0.975, 1.025)$ to break the symmetry. 
Since diffusion models also take a noisy image and time step input, we fix those to $t=\sigma_{\max} = 80.0$ and a noise input $n\sim \mathcal{N}\left(\mathbf{0}, \sigma_{\max}^2 \mathbbm{1}\right)$ that is sampled and set fixed at model initialization, before training. The selection scores are produced by $\psi_1$, which receives the features extracted by $f_{\phi}$ as input. It uses a DCGAN decoder architecture, but is trained from scratch.

For Cityscapes, we use a different split than WSS. Therefore, we retrain the Cityscapes diffusion model with our alternate split, using $\mu_{\mathrm{train}}=1.5$ as the result of hyperparameter selection. This diffusion model is then taken as the starting point for our mode proposal models. The metrics reported in \cref{tab:main-results} for the Cityscapes diffusion model are taken directly from WSS\cite{gerard_wildfire_2025}, not from our retrained model. 

\subsection{Hungarian matching}

During training, $l_{\textrm{mask}}$ in \cref{eq:optimal-matching,eq:lmask} is not the IoU, but the respective loss function used by the model, to avoid having to compute both IoU and loss for all image pairs. This is also used for model selection on the validation set during training. When evaluating the model afterwards, we always use IoU for matching, to have a consistent evaluation method across all models. 

\subsection{Hyperparameters}

Unless specified otherwise, we choose the number of proposals to be twice the number of ground truth masks provided by the respective dataset, \ie $\hat{K}_{\textrm{MMFire}} = 16, \hat{K}_{\textrm{Cityscapes}} = 32, \hat{K}_{\textrm{LIDC}} = 8$. For the single-label scenario, we use the positive-unlabeled loss with the expected relative frequency of positive samples $\eta_p=0.5$. When increasing the number of proposals in \cref{tab:redundant-proposals}, we decrease $\eta_p$ accordingly. 

The hyperparameters for training mode proposal models were chosen based on short grid searches, one for each dataset and choice of single-label or fully-labeled scenario. As loss functions for the proposals, we considered the mean squared error (MSE, as used in the diffusion models that we use as the starting point for our proposal models), Dice and a 1:1 combination of Dice and Focal loss ($\gamma=2$, no per-class prior $\alpha$). The best-performing configuration was chosen based on the achieved HM IoU* on the validation set. \cref{tab:hyperparams} shows the chosen hyperparameters. 

\begin{table}[]
    \centering
    \caption{\textbf{Training hyperparameters}, mostly determined via hyperparameter search. The number of proposals is instead set to be twice the number of ground truth masks that each dataset provides.}
    \begin{tabular}{lccllcc}
    \toprule
    Dataset & Epochs& Proposals  & Scenario & Loss function & Learning rate & $\lambda$ \\\midrule
     MMFire & 100 & 16  & Single & MSE & 1e-3 & 1e-2 \\
     MMFire & 100 & 16  & Full & DiceFocal & 1e-3 & 1e-2 \\     
     Cityscapes & 400 & 32 &Single & MSE & 1e-4& 1e-1 \\     
     Cityscapes & 400 & 32 &Full & MSE & 1e-4& 1e-1 \\     
     LIDC & 100 & 8 & Single & DiceFocal & 1e-3 & 1e-3 \\     
     LIDC & 100 & 8 & Full & DiceFocal & 1e-3 & 1e-1 \\     
    \bottomrule
    \end{tabular}
    \label{tab:hyperparams}
\end{table}

\section{Further Experimental Results}

\cref{tab:freeze-weights} shows that diffusion-pretraining is not necessary to achieve a high mode coverage, but that freezing the pretrained encoder seems to help with achieving a good selection score. Consequently, we use this setup for all of our experiments. 

\begin{table}
\centering
\caption{\textbf{Training with frozen weights - fully-labeled:} The proposal model is initialized with the pretrained diffusion model weights. Varying parts of the model are kept frozen during training. We find that preserving the diffusion-pretrained weights in the encoder leads to greatly increased selection performance .}
\label{tab:freeze-weights}
\begin{tabular}{l c c c c}
\toprule
Training regime &  Pretrained~~ & Frozen weights~~ & HM IoU*~~ & Selection F1\\
\midrule
From scratch & \xmark & \xmark &\textbf{0.927} & 0.948 \\
All weights & \cmark & \xmark & \textbf{0.927} & 0.952 \\
Decoder only & \cmark & \cmark & 0.923 & \textbf{0.994}\\
Last conv. only& \cmark & \cmark & 0.752 & 0.933 \\
\bottomrule
\end{tabular}
\end{table}

\begin{sidewaysfigure}
    \centering
    \includegraphics[width=\linewidth]{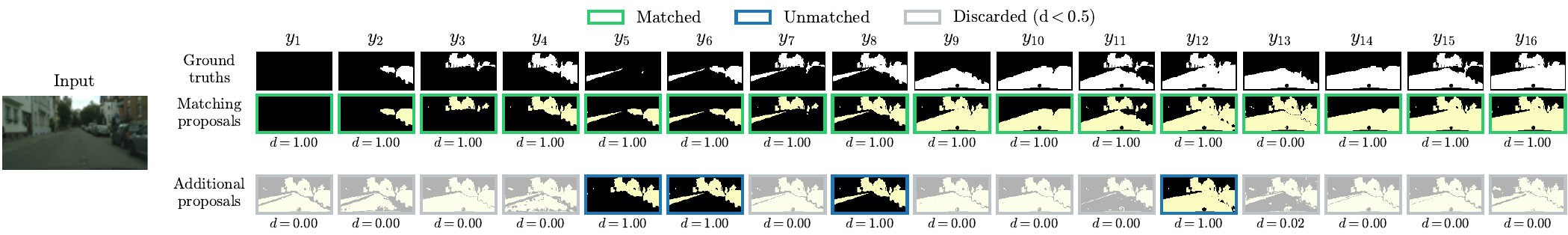}
    \caption{\textbf{Cityscapes proposals:} The binary ambiguous Cityscapes dataset creates diverse modes for a given input image by flipping each of four classes independently either to the positive or negative class, following fixed probabilities. The model used here was trained in the single-label scenario. Proposals displayed are thresholded. The model can confidently discard many redundant proposals, because the non-thresholded value ranges of those proposals are unusual, indicating that they have not been trained well. Given that only a single proposal per input is updated at each training step, this is unsurprising. Some modes correctly specialize and will often be the best-matching ones and thus receive further gradient updates, while others are too far from the correct outputs to be chosen in Hungarian matching, and thus barely receive gradient updates.}
    \label{fig:cs-proposals}
\end{sidewaysfigure}


\subsection{LIDC - Training with and without Selection Loss}

Given the large gap of $>20\%$ in standard HM IoU between FlowSSN and our results with 8 samples (\cref{tab:main-results-lidc}) on LIDC, as well as the low selection F1 scores of $<60\%$, we investigate whether the selection loss might negatively impact our model's performance. An F1 score near 50\% could indicate that the selection task is ill-posed for this dataset, likely due to the small differences between modes that make it hard to distinguish between correct and incorrect proposals. Forcing the model to optimize the corresponding loss would likely add gradient noise that might impede the optimization of the proposals.

\textbf{Selection Loss vs. HM IoU} The results in \cref{tab:lidc-no-lambda} confirm our suspicion that including the selection loss ($\lambda > 0$) negatively impacts HM IoU, especially for proposals $\hat{K} \geq 32$. However, both with and without the selection loss, the standard HM IoU grows strongly as the number of proposals grows. In contrast, the two metrics focused on \textit{unique} ground truth masks (HM IoU$_{\text{multi}}$ and HM IoU*) grow very slowly as  proposals increase. We interpret this as evidence that the model already covers the principal modes of the underlying distribution quite well with only 8 proposals. Adding more proposals mostly seems to add redundant proposals that can then be matched with duplicate proposals and thus greatly increase the HM IoU, without increasing the captured aleatoric uncertainty. Thus we chose to include several models trained with $\lambda=0$ in the comparison in \cref{tab:main-results-lidc}.

\textbf{Model Selection and Shorter Training} Our training objective does not encourage the production of duplicate proposals in any way. Thus, the model selection during training, based on validation loss, is not necessarily good at selecting a model that optimizes the standard HM IoU. By observing the validation HM IoU during training, we found that the maximum performance is achieved during the first few epochs, while further training lead to the HM IoU converging towards a lower value. To illustrate the difference: model checkpoints selected for $\hat{K}\in \{16,32\}$ proposals, when trained over 100 epochs with $\lambda=0$, achieved only 65.9\% HM IoU (-3.1\%) and 73.8\% (-5.9\%), with values in brackets indicating the distance from their 10-epoch counterparts. Thus, the results in \cref{tab:lidc-no-lambda} and \cref{tab:main-results-lidc} for $\lambda=0$ are based on training for 10 epochs instead of 100.

\begin{table}[]
    \centering
    \caption{\textbf{LIDC validation: Selection loss reduces HM IoU.} We vary the number of proposals, training with selection loss ($\lambda > 0$) or without ($\lambda=0$). HM IoU is the standard metric, matching against the full set of ground truth masks. HM IoU$_{\mathrm{multi}}$ modifies this by matching all duplicate ground truth masks to the same proposal. HM IoU* removes duplicate ground truth masks, matching proposals with the set of unique ground truths. Models were trained using only 10 epochs instead of 100.}
    \label{tab:lidc-no-lambda}
\begin{tabular}{rrrrr}
\toprule
$\lambda$ & Proposals & HM IoU & HM IoU$_{\mathrm{multi}}$ & HM IoU* \\\midrule
0.0 & 8 & 0.624 & 0.834 & 0.799 \\
0.0 & 16 & 0.690 & 0.852 & 0.818 \\
0.0 & 32 & 0.797 & 0.863 & 0.832 \\
0.0 & 64 & 0.868 & \textbf{0.872} & \textbf{0.844} \\
0.0 & 128 & \textbf{0.869} & 0.869 & 0.840 \\\midrule
0.1 & 8 & 0.622 & 0.831 & 0.794 \\
0.1 & 16 & 0.676 & 0.851 & 0.818 \\
0.1 & 32 & 0.712 & 0.849 & 0.815 \\
0.1 & 64 & 0.686 & 0.833 & 0.796 \\
0.1 & 128 & 0.815 & 0.859 & 0.828 \\
\bottomrule
\end{tabular}
\end{table}

\subsection{Estimating Class-Flip Probabilities on Cityscapes}
\label{sec:supp-cs-estimate-probabilities}

Decomposing the velocity over the full set of Cityscapes ground truths results in a relatively low Brier Skill Score of 0.443 (\cref{tab:velocity-decomp-gt-masks}). This is caused by the specific mode structure underlying the ambiguous Cityscapes dataset. The modes of this dataset are constructed as the Cartesian product of four binary decisions: each of the four classes \textit{road}, \textit{sidewalk}, \textit{vegetation}, and \textit{car} are independently flipped to either the positive or negative class following a per-class Bernoulli distribution with probabilities $\{p_{ci}\}_{1\leq i\leq4}$ (see \cref{fig:cs-proposals} for an example set of modes). 

Let $\alpha_i~\in\{0,1\}$ be the outcome of that decision for class $i$, and $y_{ci}$ the mask in which only the pixels belonging to class $i$ are set to positive (\eg all \textit{road} pixels are positive, all others are negative). We call $y_{ci}$ the \textit{template} for class $i$. $y_{ci}$ indicates where class $i$ \textit{could} appear, while $\alpha_{i}$ decides whether it \textit{does} appear. A mode $y_j$ with associated flip decisions $\{\alpha_{ij}\}_i$ would then be computed as:
\begin{equation}
    y_j = \sum_{i=1}^4 \alpha_{ij} y_{ci},
\end{equation}
which results in a binary mask, since all pixels are exclusively assigned to one class and each pixel only takes binary values. Iterating over all possible combinations of Bernoulli outcomes results in $2^4=16$ modes, including the completely empty image when $\alpha_i=0$ for all $i$. Formally, all of the mode-specific variables mentioned above should also be conditioned on the input image $x$. In this section, we choose to omit this additional index for increased readability.

As a result of this specific mode structure, the correct mode distribution is not the unique solution to the constrained least squares problem in ~\cref{eq:pi_optimization}, which explains the BSS of 0.443. However, by integrating knowledge about this lower-dimensional structure in the numerical optimization, decomposing the problem further into per-class flip probabilities, we recover the $p_{ci}$ via velocity decomposition and achieve a high BSS of 0.870 over the full validation set (measured against the class-flip probabilities, not per-mode probabilities; see below for details).

Our general idea is to take a hierarchical probability estimation approach:  First, we compute a solution to the underconstrained least squares problem. Then, we use that solution as part of a second optimization problem, which uses the knowledge of the class-flipping structure, to find the Bernoulli probabilities $p_{c}$. 

The ground truth probability of mode $y_j$ resulting from the class-flip probabilities is computed as follows:

\begin{equation}
\label{eq:cs-gt-probabilities}
    p_1(y_j \mid x, p_{c}) = \prod_{i=1}^4 p_{ci}^{\alpha_{ij}} (1-p_{ci})^{(1-\alpha_{ij})}.
\end{equation}

These are exactly the probabilities that the initial least squares problem in \cref{eq:pi_optimization} tries to estimate. Let $\hat{w}$ be a solution found by this initial velocity decomposition. If we assume that for a set of $\{y_j\}_j$ the $\{\alpha_{ij}\}_{ij}$ are known, we can then take $\hat{w}$ as fixed target values and approximate it with \cref{eq:cs-gt-probabilities}, iterating over values for $p_c$ to estimate class-flip probabilities that best explain the $\hat{w}$. We measure the approximation error via KL divergence:

\begin{align}
    &D_{\mathrm{KL}} \left( \hat{w} \mid\mid p_1\left(y \mid x, \hat{p}_{c}\right)  \right)\\
    &= \sum_j  \hat{w}_j \log\left( \frac{\hat{w}_j}{p_1\left(y \mid x, \hat{p}_{c}\right)} \right)\\
    &= \sum_j  \hat{w}_j \log\hat{w}_j - \sum_j  \hat{w}_j \log p_1\left(y \mid x, \hat{p}_{c}\right).
\end{align}

In contrast to the initial least squares problem, the desired probabilities $\{\hat{p}_{ci}\}_i$ are now independent Bernoulli probabilities. Thus, they are not constrained to sum to one, but only required to be non-negative. When minimizing this term, the first sum is constant wrt. $\hat{p}_{c}$ and can therefore be omitted:

\begin{align}
    &\arg\min\limits_{\hat{p}_{c}} D_{\mathrm{KL}} \left( \hat{w} \mid\mid p_1\left(y \mid x, \hat{p}_{c}\right)  \right)\\
    = & \arg\min\limits_{\hat{p}_{c}}  \underbrace{\sum_j  \hat{w}_j \log\hat{w}_j}_{\text{const.}} - \sum_j  \hat{w}_j \log p_1\left(y \mid x, \hat{p}_{c}\right) \\
    = & \arg\min\limits_{\hat{p}_{c}}  - \sum_j  \hat{w}_j \log p_1\left(y_j \mid x, \hat{p}_{c}\right) \\
    = & \arg\min\limits_{\hat{p}_{c}}  - \sum_j  \hat{w}_j \log\left(\prod_{i=1}^4 \hat{p}_{ci}^{\alpha_{ij}} (1-\hat{p}_{ci})^{(1-\alpha_{ij})}\right) .
    \label{eq:cs-estimate-class-probabilities}
\end{align}

This minimization problem can be solved numerically, provided we have access to $\{\alpha_{ij}\}_{ij}$. When using the set of ground truth masks as $\{y_j\}_j$, we know $\{\alpha_{ij}\}_{ij}$ by construction. The resulting estimated per-class probabilities $\hat{p}_c$ achieve a high BSS of 0.870 on our Cityscapes validation set. Note that this BSS is now measured against the per-class flip probabilities, while the BSS of 0.443 was measured against the per-mode probabilities. This improved approximation result proves that velocity decomposition can also successfully be used on this more challenging dataset, if the underlying structure is known and can be integrated in the optimization procedure. Next, we show why moving beyond this baseline computed with access to the ground truths is difficult. 

\subsubsection{Velocity decomposition using proposals} When using proposals instead of ground truth masks, we can still use \cref{eq:pi_optimization} to compute $\hat{w}$. However, the flip decisions $\{\alpha_{ij}\}_{ij}$ are not available anymore, so \cref{eq:cs-estimate-class-probabilities} can not be evaluated right away. If we had access to the class templates $y_{ci}$, these $\alpha$ values could be estimated by thresholding the overlap between proposals $y_j$ and templates $y_{ci}$:

\begin{equation}
    \label{eq:estimate-alpha}
    \hat{\alpha}_{ij} = \mathbbm{1}(\frac{1}{L}\sum_{l=1}^Ly_{j}^l\cdot y^l_{ci} > 0.5),
\end{equation}
where $\mathbbm{1}$ is the indicator function and the index $l$ iterates over the $L$ pixels of the respective mask. With this, the class $i$ is estimated to be 'flipped on' if at least half of its pixels are 'on' in the proposal.

To estimate the required class templates $y_{ci}$ from proposals, we now take the following approach: We compute a principal component analysis (PCA) of the set of proposals, and threshold the four first principal components to binary masks. These serve as the class templates. Because the ground truth modes only differ by flipping the whole area associated with a class $i$, exactly these areas should emerge as the main directions of variation that the PCA computes. With access to these estimated binary class templates, we can then estimate $\{\alpha_{ij}\}_{ij}$ for the proposals $\{y_j\}_j$ via \cref{eq:estimate-alpha}, and finally estimate the class-flip probabilities $\hat{p}_c$ via \cref{eq:cs-estimate-class-probabilities}.

\begin{figure}
    \centering
    \includegraphics[width=\linewidth]{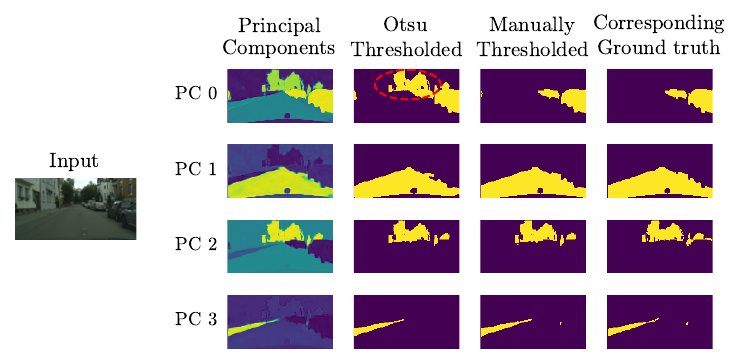}
    \caption{\textbf{Cityscapes: Estimating class templates from proposals} First, proposals are filtered according to their selection scores (for this picture: from 32 to 22, not pictured). Second, principal components are computed based on these proposals (first column, one row per component) and then thresholded into binary masks that are supposed to represent one class each (second and third column). Otsu's method does not separate the vegetation and car class in the first component (marked by the dashed ellipse), making it impossible to correctly estimate the underlying probabilities, though this is possible with a manually chosen threshold.}
    \label{fig:cs-decomposition-pca-results}
\end{figure}

\cref{fig:cs-decomposition-pca-results} shows the resulting first four principal components for an example image. Automatic thresholding via Otsu's method~\cite{otsu_threshold_1979} fails to separate the vegetation and car classes, preventing us from correctly estimating the $\alpha_{ij}$. 
However, by applying manual thresholding, we are able to separate all four classes and compute sensible $\hat{\alpha}_{ij}$ for each proposal $j$ and class $i$. Subsequently, we compute $\hat{p}_c$, achieving a BSS of 0.983 for this image. \cref{tab:cs-manual-thresholding-probabilities} compares the estimated probabilities for this image, using either manually thresholded principal components as class templates, or the ground truth masks.

While the need for manual thresholding makes an evaluation for the whole validation set infeasible, the successful prior estimation is a proof of concept. It demonstrates the principal utility of the velocity decomposition approach, even for special cases in which the dataset has a complex underlying structure.

\begin{table}[]
    \centering
    \caption{\textbf{Cityscapes: Estimated probabilities} for the image in \cref{fig:cs-decomposition-pca-results}. Both the manually thresholded principal components, computed from proposals, as well as the ground truth masks, lead to very good estimates of the target distribution, though both underestimate the sidewalk class by about 0.11.}
    \label{tab:cs-manual-thresholding-probabilities}
    \begin{tabular}{lccccc}\toprule
         Method & $p_{\mathrm{road}}$ & $p_{\mathrm{sidewalk}}$ & $p_{\mathrm{vegetation}}$ & $p_{\mathrm{car}}$ & BSS\\\midrule
        Target distribution & 0.050& 0.250 & 0.750 & 0.950 & - \\
        Proposals (manual) & 0.050 & 0.142 & 0.762 & 0.910 & \textbf{0.983} \\
        GT masks & 0.053 & 0.145 & 0.706 & 0.921 & 0.982 \\\bottomrule
    \end{tabular}
    
\end{table}

\subsection{Varying the Number of Proposals}
\label{sec:vary-number-of-proposals}

Since the mode proposal models are deterministic convolutional networks, we have to choose $\hat{K}$, the number of proposals to output, at training time. For this paper, we always choose twice the number of ground truth masks that the respective dataset provides, unless indicated otherwise. However, in a practical setting, this number would be unknown, such that we might vastly over-estimate the required number of proposals. \cref{tab:redundant-proposals} luckily shows that even having three times more proposals than necessary does not degrade segmentation performance. For the single-labeled setting, increasing $\hat{K}$ from the ground truth 8 to 32 even leads to a gain of $3.1\%$ HM IoU*. The fully-labeled scenario always matches the ground truth masks to the $K_x$ best proposals, and thus allows for parallel specialization of proposals. In contrast, when only one proposal is updated at each point, multiple different ground truth masks might be matched to the same proposal. Increasing the number of proposals to choose from makes it more likely that different ground truth masks are matched to different proposals, thus reducing conflicting training signals and improving performance.

In \cref{tab:redundant-proposals}, we see that the selection F1 score is not impacted in the fully-labeled scenario. However, in the single-label scenario, the positive-unlabeled loss does not seem to have enough information to train the selection scores if the number of proposals becomes too big. On the other hand, it can also be seen as a success that we reach $69.2\%$ F1 Score when only a single target out of 16 is labeled, and several modes occur with less than 1\% probability. Unlike for the proposals, all selection scores are updated at each step.

\begin{table}[t]
\centering
\caption{\textbf{Ablation - Redundant proposals, fully labeled:} We  investigate how different choices for the number of proposals $\hat{K}$ influences the model performance. Note that the selection scores for eight proposals are trivial (marked in italic), since the number of proposals equals the number of ground truth masks, unless duplicate ground truth masks occur. Dataset: MMFire validation}
\label{tab:redundant-proposals}
\begin{tabular}{l c c c}
\toprule
Training targets ~~& Proposals ~~&  HM IoU*~~ & Selection F1\\
\midrule
Full & 8 &  0.923 & \textbf{\textit{0.997}}\\
Full & 16 & 0.923 & 0.994\\
Full & 32 & \textbf{0.924} & 0.978\\\midrule

Single & 8 & 0.883 & \textbf{\textit{0.995}} \\ 
Single & 16 & 0.910 & 0.692 \\
Single & 32 & \textbf{0.914} & 0.306 \\

\bottomrule
\end{tabular}
\end{table}

\section{Monte Carlo Samples in Velocity Decomposition}

When estimating the mode probabilities via velocity decomposition, we have to choose a number of prior noise samples for which to compute the velocity, which are then decomposed into a weighted sum via a numerical solver. In \cref{tab:vary-mc-samples}, we see that increasing the number of samples increases the Brier Skill Score. We use 64 samples for all of our experiments. More samples might provide slightly higher scores still, but also increase computation times further, since each sample requires an evaluation of the flow model. 

\begin{table}[t]
\centering
\caption{\textbf{Ablation: Vary number of Monte Carlo samples in mode probability estimation.} We vary the number of initial samples $y_t$ to approximate the expectation in \cref{eq:pi_optimization}.}
\label{tab:vary-mc-samples}
    \begin{tabular}{r r}
    \toprule
    Samples ~~& Brier Skill Score \\\midrule
    8 & 0.913 \\
    16 & 0.910 \\
    32 & 0.916 \\
    64 & \textbf{0.919} \\
    \bottomrule
    \end{tabular}
\end{table}

\end{document}